\documentclass[fleqn,10pt]{wlscirep}
\usepackage[utf8]{inputenc}
\usepackage[T1]{fontenc}
\usepackage{lineno}
\usepackage{enumitem}
\usepackage{forest}
\usepackage{dirtree}
\usepackage{setspace}

\usepackage[most]{tcolorbox}
\usepackage{lipsum}  % This package generates filler text.

\def\OURDATA{\textbf{RadGenome-Chest CT}}

\title{
RadGenome-Chest CT: A Grounded Vision-Language Dataset for Chest CT Analysis.}

\author[1,2]{Xiaoman Zhang}
\author[1,2]{Chaoyi Wu}
\author[1,2]{Ziheng Zhao}
\author[2,3]{Jiayu Lei}
\author[1,2]{\\ \vspace{2pt} Ya Zhang} 
\author[1,2, $\dagger$]{Yanfeng Wang}
\author[1, 2, $\dagger$]{Weidi Xie}
\affil[1]{Shanghai Jiao Tong University, Shanghai, China}
\affil[2]{Shanghai AI Laboratory, Shanghai, China}
\affil[2]{University of Science and Technology of China, Anhui, China}

\affil[$\dagger$]{Corresponding Authors}

\begin{abstract}

Developing generalist foundation model has recently attracted tremendous attention among researchers in the field of AI for Medicine~(AI4Medicine). A pivotal insight in developing these models is their reliance on dataset scaling, which emphasizes the requirements on developing open-source medical image datasets that incorporate diverse supervision signals across various imaging modalities. In this paper, we introduce \OURDATA, a comprehensive, large-scale, region-guided 3D chest CT interpretation dataset based on CT-RATE. Specifically, we leverage the latest powerful universal segmentation and large language models, to extend the original datasets~(over \textbf{25,692} non-contrast 3D chest CT volume and reports from \textbf{20,000} patients) from the following aspects: 
(\romannumeral1) organ-level segmentation masks covering \textbf{197} categories, which provide intermediate reasoning visual clues for interpretation;
(\romannumeral2) \textbf{665 K} multi-granularity grounded reports, where each sentence of the report is linked to the corresponding anatomical region of CT volume in the form of a segmentation mask; 
(\romannumeral3) \textbf{1.3 M} grounded VQA pairs, where questions and answers are all linked with reference segmentation masks, enabling models to associate visual evidence with textual explanations. All grounded reports and VQA pairs in the validation set have gone through manual verification to ensure dataset quality. We believe that \OURDATA~ can significantly advance the development of multimodal medical foundation models,
by training to generate texts based on given segmentation regions, which is unattainable with previous relevant datasets. 
We will release all segmentation masks, grounded reports, and VQA pairs to facilitate further research and development in this field.

\end{abstract}
\begin{document}

\flushbottom
\maketitle
%  Click the title above to edit the author's information and abstract

% \thispagestyle{empty}

% \noindent Please note: Abbreviations should be introduced at the first mention in the main text – no abbreviations lists or tables should be included. The structure of the main text is provided below.

\section*{Background \& Summary}

% \textbf{\xiaoman{Why do we need this dataset?}}
% The medical domain lacks extensive datasets that provide deep annotations linking visual features with medical knowledge and patient data. Grounded datasets in medical imaging would allow for the development of algorithms that reflect the diagnostic reasoning of radiology experts more closely. This could enhance AI's ability to mimic human experts' decision-making processes, improving its utility in clinical settings.

% By proposing these datasets, researchers aim to develop AI systems that can not only recognize and label medical images accurately but also understand and describe complex relationships between localized features within these images. This is akin to how radiologists interpret images, considering not just the visual data but also its clinical significance in the broader context of patient care.

% \xiaoman{Will add more citations later}
In the recent literature, 
the evolution of large-scale foundation models~\cite{ lu2023towards,achiam2023gpt,singhal2023large,team2023gemini,alayrac2022flamingo} has sparked significant interest in the development of generalist medical AI~(GMAI) systems~\cite{wu2023towards,tu2024towards,li2024llava,huang2023visual,moor2023med}, particularly within the realm of radiology—a crucial component of medical diagnostics. By training on large-scale visual-language medical datasets, {\em i.e.}, medical scans paired with global clinical reports, for example, MIMIC-CXR~\cite{johnson2019mimic} has chest X-ray scans from 227,835 studies, and CT-RATE~\cite{hamamci2024foundation} contains chest CT scans from 20,000 patients. These medical models have demonstrated the preliminary ability for writing clinical reports,
aiming to support radiologists throughout their workflow and markedly reducing workloads.

However, existing datasets only provide global reports for the medical scan, which has posed limitations on training models that
enables grounded report generation, grounded question answering,
{\em i.e.}, describing regional abnormalities and relevant normal findings, or answer questions corresponding to certain regions.
To further push forward the training of more capable generalist models, we propose to extend the existing image-reports datasets with region-wise description, {\em i.e.}, linking the descriptive labels or findings from diagnostic reports to their corresponding anatomical regions in the images, in the form of segmentation masks for explainability. 
 
%However, due to the privacy and annotation cost in medicine, data scarcity is greatly hindering the development of GMAI, which heavily relies on training data scaling~\cite{kaplan2020scaling,brown2020language}. 

\vspace{2pt}
In this paper, we introduce \OURDATA, a comprehensive, large-scale and fine-grained annotated dataset for 3D chest CT interpretation, built upon the publicly available CT-RATE~\cite{hamamci2024foundation}. {\em Initially}, we employ the latest powerful text-prompted universal segmentation model, SAT~\cite{zhao2023one}, to segment primary anatomical targets in the image. {\em Subsequently}, utilizing large language models and NER models, we break all reports into an anatomically hierarchical structured format, and link the reports' sentences to visual regions in CT volume. {\em Finally}, we further generate visual question-answering pairs closely related to the structured report and segmented image, from both region level and case level. In summary, we have extended the original image-report datasets from the following aspects:

\vspace{-3pt}
\begin{enumerate}[label=\roman*)]
    \setlength\itemsep{1pt}
    \item Organ-level segmentation masks that covers \textbf{197} categories, {\em i.e.}, all the critical regions existing in clinical CT reports;
    \item \textbf{665k} multi-granularity grounded reports, with each sentence grounded to the corresponding anatomical region.
    \item \textbf{1.3 M} grounded VQA pairs, concerning both critical region-wise findings and comprehensive case-wise impressions. All the questions and answers are linked to segmentation masks for reference. % enabling models to associate visual evidence with textual explanations. 
\end{enumerate}

\noindent We believe that \OURDATA~, 
with provided region-to-report associations,
can significantly advance the development of agent-based multimodal medical foundation models, that enables to generate texts, grounded on the corresponding visual regions, which is unattainable with previous relevant datasets.

\begin{figure}[tb]
\centering
\includegraphics[width=\linewidth]{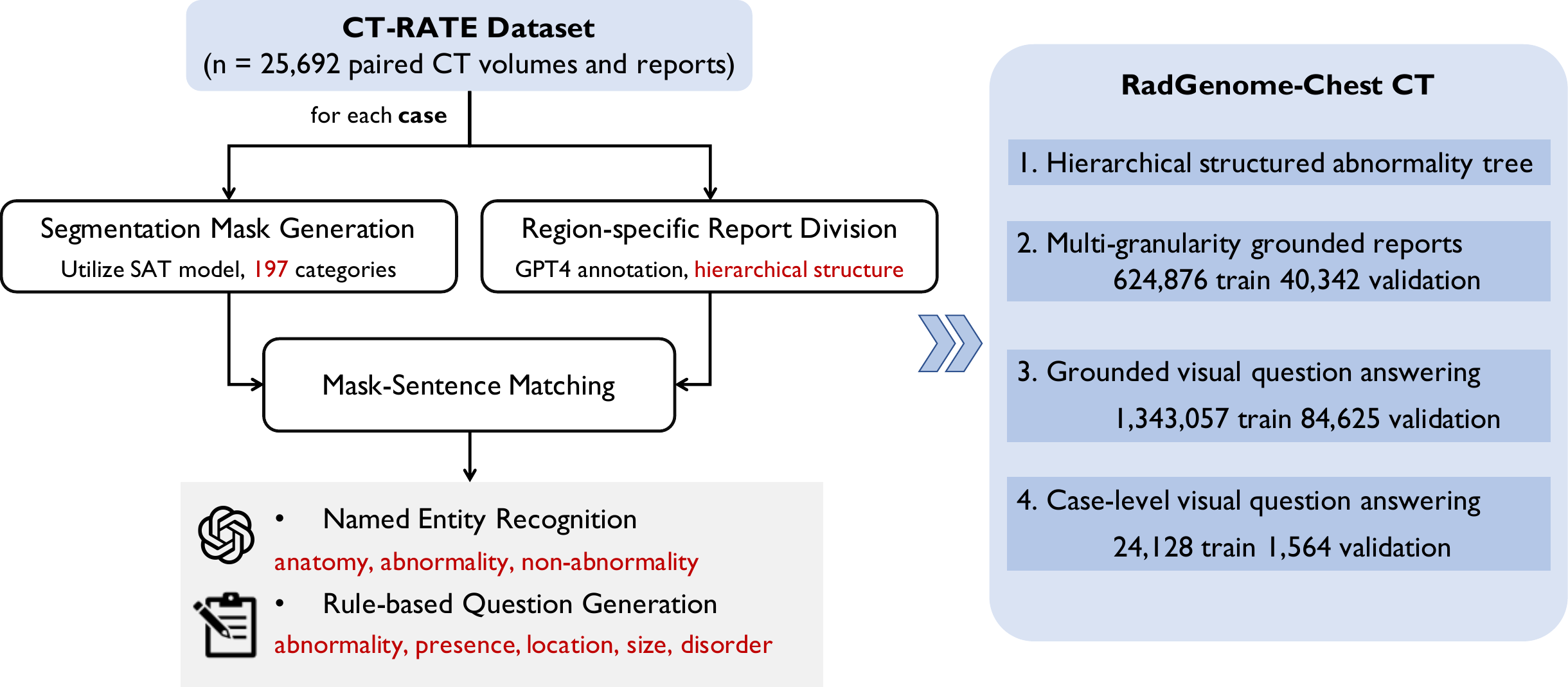}
\caption{Data construction pipeline of \OURDATA.
% \weidi{apart from this pipeline figure, can we add an example figure, 
% that can sort of showing the results from each step.}\xiaoman{update later}
}
\label{fig:dataset_pipeline}
\end{figure}

\section*{Methods}
% \xiaoman{detailed text describing any steps or procedures including full descriptions of the experimental design, data acquisition assays, and any computational processing (e.g. normalization, image feature extraction). There is no limit to the length of the Methods section. Subheadings should not be numbered.}

\begin{figure}[!t]
\centering
\includegraphics[width=\linewidth]{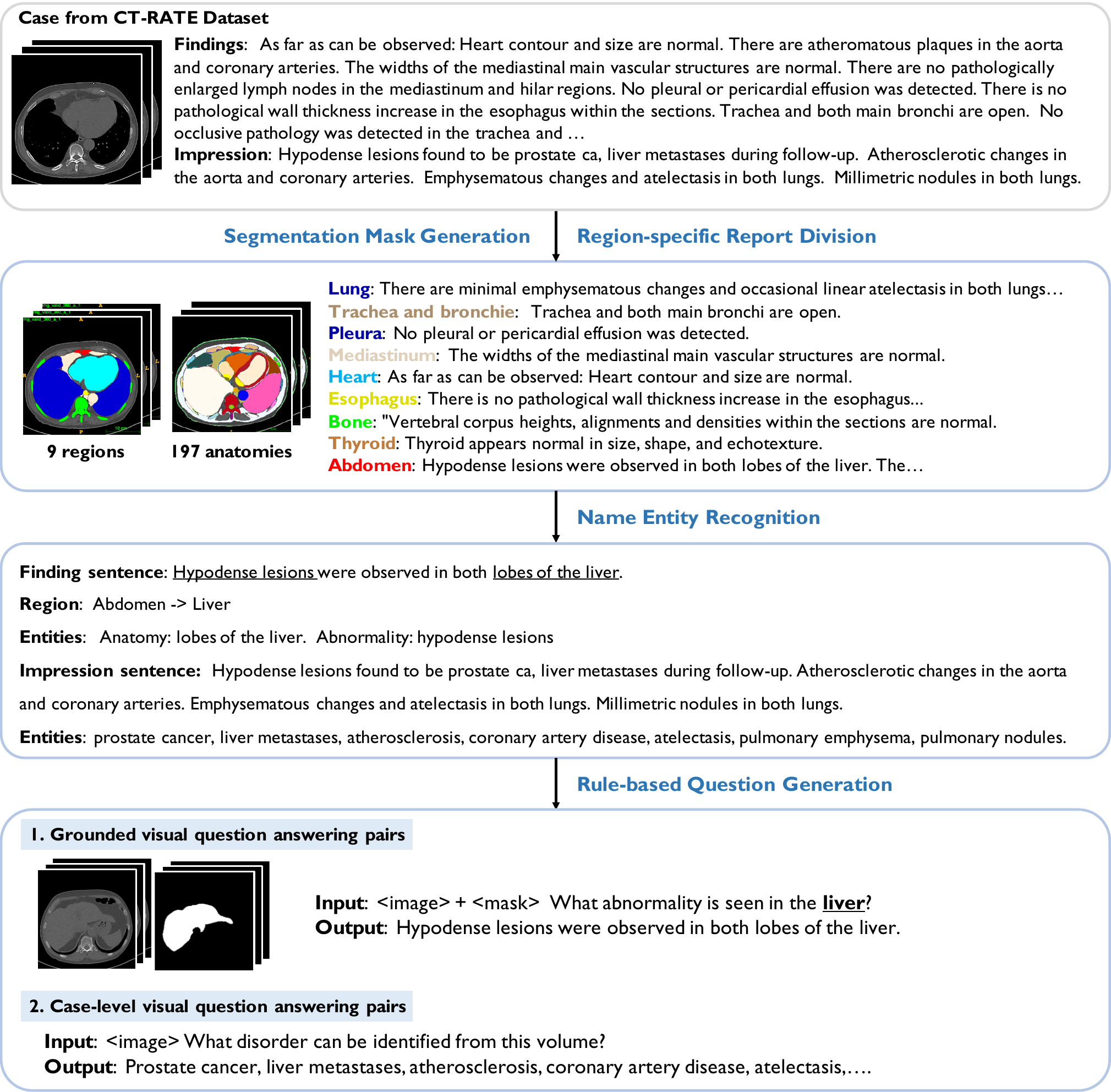}
\caption{Overview of results obtained from each step of the data construction pipeline.}
\label{fig:dataset_overview}
\end{figure}

In this section, we start by introducing the source dataset that \OURDATA~is built on.
Next, we provide a detailed description of the collection procedure for obtaining segmentation masks, and region-wise reports, as illustrated in Fig.~\ref{fig:dataset_pipeline}.
The outcomes from each step are presented in Fig.~\ref{fig:dataset_overview}.

% Then, based on \OURDATA, we formally define the \OURDATA benchmarks and introduce the baseline models that we evaluate in this article.

\subsection*{Data Source}
We initiate our study with CT-RATE~\cite{hamamci2024foundation}~(\url{https://huggingface.co/datasets/ibrahimhamamci/CT-RATE}), 
it is a dataset of 25,692 non-contrast 3D chest CT volumes derived from 21,304 unique patients, each volume is accompanied by a radiology text report and annotated with 18 distinct types of abnormalities. These 25,692 non-contrast 3D chest CT volumes have been reconstructed with various methods to accommodate different window settings, totaling 50,188 images. For consistency in this paper, we have standardized all CT volumes to a uniform voxel spacing of $1\times1\times3 mm$, resulting in only 25,692 paired CT volumes and reports. We follow the official division: 20,000 patients~( 24,128 volumes) were allocated to training and 1,304~(1,564 volumes) for validation.

\subsection*{Constructing \OURDATA}
The pipeline consists of three major stages, as shown in Fig.~\ref{fig:dataset_pipeline}:
\textbf{(i) segmentation mask generation}, where detailed masks for each anatomical region in the chest CT volumes are created;
\textbf{(ii) region-specific report division}, that involves the annotation and categorization of radiology text reports by the anatomical regions they refer;
\textbf{(iii) rule-based question generation}, which entails extracting entities from the sentence, and formulating visual question answering (VQA) pairs linked to specific segmentation masks.

\subsubsection*{Segmentation Mask Generation} 
To segment as many anatomical regions as possible, 
we employ the recent SAT~\cite{zhao2023one} model.
It is a knowledge-enhanced segmentation model, that employs natural language as prompts to effectively segment 3D medical volumes. The model has been trained on 72 diverse segmentation datasets, covering 498 classes across various anatomical regions including the brain, head and neck, thorax, spine, abdomen, and limbs. For our research, SAT is adopted to execute detailed segmentation across all volumes of the CT-RATE dataset. 
Specifically, we focus on segmenting \textbf{197} regions pertinent to chest CT scans, enabling precise anatomical analysis.
The list of segmented anatomies was organized into a hierarchical tree as shown in Supplementary~\ref{sec:hierarchy_anatomical}.
This includes several major regions such as the lungs, trachea and bronchi, mediastinum, heart, pleura, bones, thyroid, breasts, abdomen, and other areas.
% \xiaoman{TODO: add a figure to show all regions that we provided. \textbf{(LOWER PRIORITY)}}
% \xiaoman{Furthermore, to enhance the segmentation performance of SAT, we employ the segmentation masks generated by SAT as prompts for MedSAM~\cite{}.}

\subsubsection*{Region-wise Reports Generation}
The goal here is to break the entire reports into region-wise descriptions, we observe that the OpenAI GPT-4~\cite{achiam2023gpt} model can complete such task in very high accuracy~\cite{nori2023capabilities,liu2023exploring}. However, employing GPT-4 on the entire set would be prohibitively expensive, thus, we train a model for such report division. Specifically, we first employ GPT-4~\cite{achiam2023gpt} to annotate the anatomical regions of each sentence in the ``\textbf{FINDINGS}'' section of 2,500 radiology reports, which comprise all reports in the validation set. The prompt used is as follows.
\begin{center}
    \begin{tcolorbox}[width=0.98\textwidth, colback={gray!0}, boxrule=0.3mm, arc=1mm, auto outer arc,colframe=black, center title]
        You are a radiologist tasked with extracting anatomical regions from the ``FINDINGS'' section of radiology reports. For each sentence provided, identify the corresponding anatomical regions. Ensure each identified region is an entry from a predefined list: [region\_list]. If a sentence mentions ``left'' or ``right'', these qualifiers should precede the anatomical region (e.g., left kidney). Given input in the format: <Input><findings><$\backslash$Input>. Please reply in the following JSON format: \{<sentence>: [region$_1$,region$_2$,...], <sentence>: [region$_1$]\}.
    \end{tcolorbox}
\end{center}

\noindent This process results in 15,926 annotated sentences. Subsequently, we divide these sentences into training and validation subsets in an 8:2 ratio and train a GPT-2 model using the annotated sentence along with the two preceding and following sentences from the report as input, if available. 
The model is designed to output the list of anatomical regions associated with each sentence. For instance, for the target sentence ``No pleural effusion was detected on the left.'', 
the input is ``There is minimal pleural effusion on the right. No pleural effusion was detected on the left. Atelectasis is observed in the middle lobe and lower lobe of the right lung. A malignant mass is observed around the lower lobe bronchi of the left lung.'', and the expected output is ``left lung''. 
The model achieves an accuracy of \textbf{94.56\%} on the validation set. Consequently, we employ this model to perform inference on all sentences across the entire dataset of reports.
More examples of segmentation results and structured reports can be seen in Supplementary Section~\ref{sec:supple_case}.

\subsubsection*{Named Entity Recognition}
Through the abovementioned process, each ``\textbf{FINDINGS}'' is divided into multiple sentences, and each is associated with one or multiple segmented regions.  
To facilitate the generation of detailed question-answer pairs from these sentences, we initially employ an in-house Named-Entity Recognition (NER) model to analyze all sentences.
This process involves extracting entities that can be categorized into ``anatomy'', ``abnormality'', and ``non-abnormality''. Here, ``anatomy'' pertains to the anatomical regions, ``abnormality'' refers to findings or diseases identified as present, and ``non-abnormality'' indicates findings or diseases that are reported as absent.
Subsequently, all extracted "abnormality" and "non-abnormality" entities undergo quality evaluation using GPT-4, allowing us to filter out and revise any inaccuracies. 
For instance, in cases where the NER model extracts abnormalities such as ``structural distortion and volume loss'', GPT-4 will segment it into ``structural distortion'' and ``volume loss'' for more accurate categorization. 
We filter out abnormalities with a GPT-4 output of ``no'' and update them to the revised versions provided by GPT-4.
The prompt used is as follows.
\begin{center}
    \begin{tcolorbox}[width=0.98\textwidth, colback={gray!0}, boxrule=0.3mm, arc=1mm, auto outer arc, colframe=black, center title]
       You are a radiologist. You will receive an input that consists of a sentence from the findings section of a chest CT report, followed by a phrase extracted from that sentence. Assess whether the extracted phrase is an abnormality. The abnormality should not be a size or a location. If the extracted phrase does not accurately represent an abnormality, your output should provide the correct abnormalities. The input format will be <Input><sentence><sep><phrase></Input>. Your response should be in JSON format, structured as follows: \{abnormality: [abnormality1, abnormality2,...], phrase: ``yes'' or ``no''\}. Note that the output should be in JSON format.
    \end{tcolorbox}
\end{center}

% \begin{center}
%     \begin{tcolorbox}[width=0.98\textwidth, colback={gray!0}, boxrule=0.3mm, arc=1mm, auto outer arc, colframe=black, center title]
%        You are a radiologist performing clinical term extraction from the ``FINDINGS'' section in the radiology report. Here a clinical term can be in ['anatomy','abnormality\_present','abnormality\_notpresent']. 'anatomy' refers to the anatomical body; 'abnormality\_present' refers to findings or disorders that are present according to the sentence; 'abnormality\_notpresent' refers to findings or disorders are not present according to the sentence. Given a list of radiology sentence input in the format: <Input><sentence><sentence><$\backslash$Input>. Please reply with the JSON format following template: \{<sentence>: \{entity: entity type, entity: entity type \}, <sentence>: \{entity: entity type\}\}
%     \end{tcolorbox}
% \end{center}

\noindent In addition, for the ``\textbf{IMPRESSION}'' section, we directly utilize GPT-4 to extract all disorders mentioned, and detailed information regarding the presence of any abnormalities in specific anatomical regions. The prompt used is as follows.
\begin{center}
    \begin{tcolorbox}[width=0.98\textwidth, colback={gray!0}, boxrule=0.3mm, arc=1mm, auto outer arc, colframe=black, center title]
    You are a radiologist performing disorders summary from the ``IMPRESSION'' section in the patient's chest CT scan report. Given input in the format: <Input><impression><$\backslash$Input> Please reply with the JSON format following template: \{disorders:[disorder$_1$,disorder$_2$,...]\}
    \end{tcolorbox}
\end{center}

%With such entity extraction pipeline, we have now acquired labels for each disorder identified and detailed information regarding the presence of any abnormalities in specific anatomical regions. \weidi{we use GPT4 for everything here ? then why not the previous section ? I thought it was due to the price ?}

\begin{table}[tbh]
\centering
\setlength{\tabcolsep}{5pt}
\begin{tabular}{llllll}
\toprule
Level & Question type & Answer type & Example & Train & Validation \\
\midrule
% Case & Disorders & Close/Open &  What diseases does this patient have? \\
Region & Abnormality &  Open &  What abnormality is seen in the \{region
\}? & 483,507 & 30,831 \\
Region & Presence & Close &  Is there evidence of \{abnormality\} in the \{region
\}? & 556,050 & 34,785 \\
Region & Location & Close \& Open & Where in the image is the \{abnormality\} seen? & 285,111 & 17,750 \\
Region & Size & Open & What is the size of the \{abnormality\} in the \{region\}? & 18,389 & 1,259\\
Case & Disorder & Open & What disorder can be identified from this volume? & 24,128 & 1,564\\
\bottomrule
\end{tabular}
\caption{\label{tab:datatype} Question and answer types in the proposed datasets. The placeholders \{region
\} and \{abnormality\} in the question templates are dynamically replaced with entities extracted from sentences.}
\end{table}

\begin{figure}[!t]
\centering
\includegraphics[width=\linewidth]{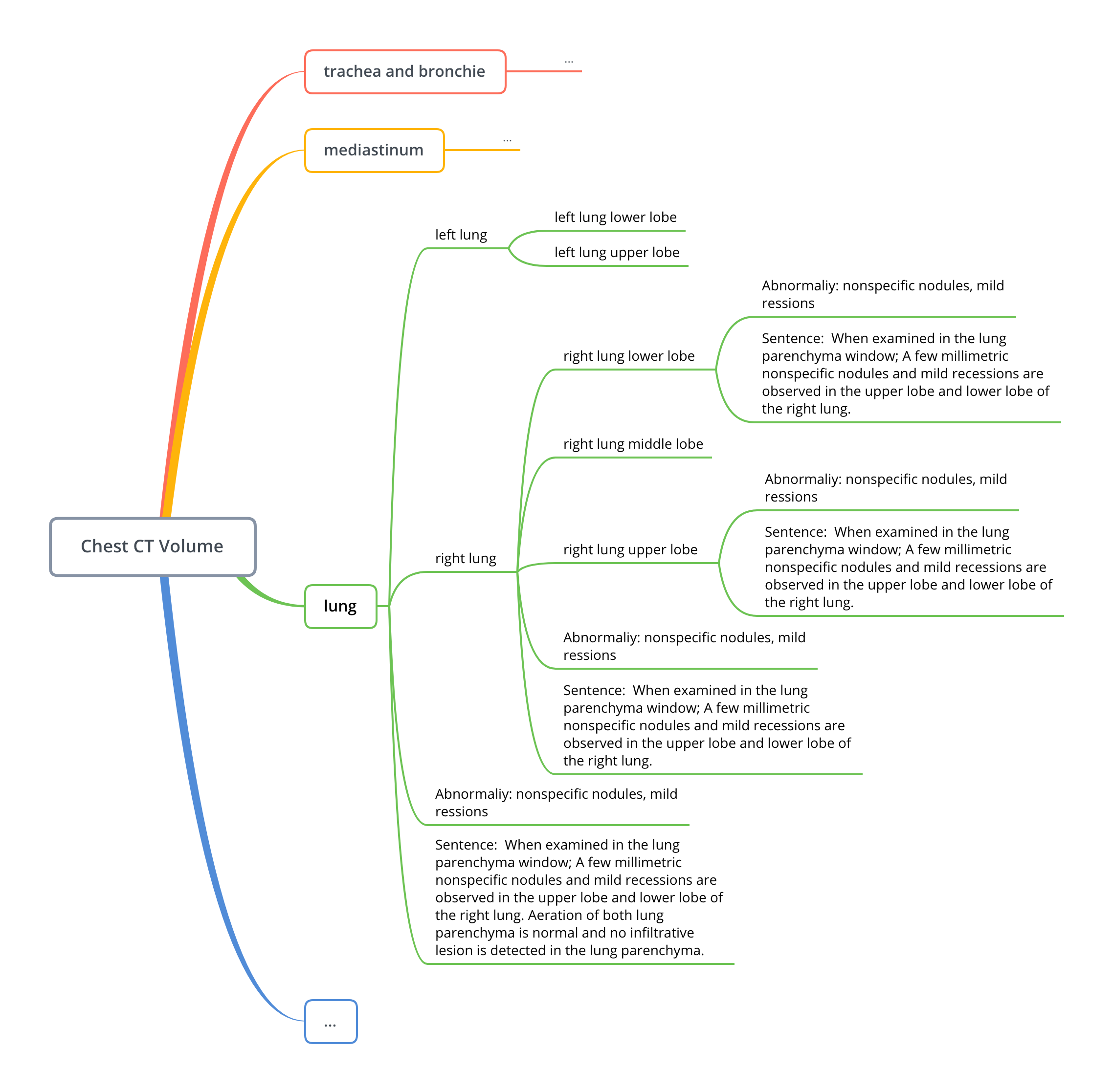}
\caption{Partial visualization of the anatomical hierarchical tree with abnormalities annotated.}
\label{fig:dataset_hierarchy}
\end{figure}

\subsubsection*{Rule-based Question Generation}
Here, we describe the procedure for generating grounded visual question-answering (VQA) data based on ``\textbf{FINDINGS}'' extracted from the report section, while case-level visual question-answering data is generated from the ``\textbf{IMPRESSION}'' section. 
Tab.~\ref{tab:datatype} presents the 5 question types in the proposed dataset.
The detailed rules will be introduced in the following sections.

\noindent First, after the region-wise report generation, 
the sentences in the findings section can be classified as follows:
\vspace{-3pt}
\begin{itemize}
    \item \textbf{Normal Findings:} Sentences that report no significant changes from normal health conditions.
    \vspace{-5pt}
    \begin{itemize}[itemsep=-1pt]
        \item \textbf{No abnormality entities} in the sentence: Sentences that mention specific anatomical regions without noting any abnormalities. For example, ``Thoracic aorta diameter is normal.''
        \item \textbf{No anatomy entities} in the sentence: Sentences that solely note the absence of specific abnormalities without referring to any anatomical regions. For example, ``No pleural effusion was detected.''
        \item \textbf{With anatomy and abnormality entities} in the sentence: Sentences that explicitly state the absence of abnormalities in specific anatomical regions. For example, ``Bilateral adrenal glands were normal and no space-occupying lesion was detected.''
    \end{itemize}
    \vspace{-8pt}
    \item \textbf{Abnormal Findings:} Sentences that report differences from normal anatomical conditions.
    \vspace{-5pt}
    \begin{itemize}[itemsep=-1pt]
        \item \textbf{No anatomy entities} in the sentence: Sentences that report an abnormal finding, 
        but do not specify an anatomical region. For example, ``Mild hiatal hernia is observed.''
        \item \textbf{With anatomy entities} in the sentence: Sentences that include both an anatomical reference and describe an abnormality. For example, ``There is narrowing of the spinal canal at the dorso-lumbar level.''
    \end{itemize}
\end{itemize}

We then construct an anatomical disorder tree for each report, based on the anatomical hierarchical tree introduced in Supplementary Section~\ref{sec:hierarchy_anatomical}. 
This involves marking any abnormalities on the tree for all nodes within the hierarchy if they are present.
As shown in Fig.~\ref{fig:dataset_hierarchy}
This comprehensive data enables us to transit to generating questions.
Taking inspiration from the previous research~\cite{hu2023expert,lau2018dataset},
we categorize the questions into four types: 
1) abnormality, 2) presence, 3) location, 4) size. Tab.~\ref{tab:datatype} shows examples of the different question types. Note that we have designed 50 templates for each question type. The details of all templates are provided in the supplementary materials. 
% \xiaoman{the templates need to be checked and polished later}
% \ziheng{cite if the QA generation refers to other paper}
%Questions are generated from these templates, specifically tailored to correspond with the type of findings described in the text.
For instance, when analyzing a sentence indicating normal findings, such as  ``Bilateral adrenal glands were normal and no space-occupying lesion was detected.'', questions can include  ``Is there any evidence of abnormality in adrenal glands?'' and ``What abnormality is seen in the adrenal glands?''.
Conversely, for a sentence in abnormal findings, ``There is a narrowing of the spinal canal at the dorso-lumbar level.'',  the question can be ``Is there any evidence of narrowing in the spinal canal?'' ``What abnormality is seen at the dorso-lumbar level of the spinal canal?'' and ``Where in the spinal canal is the narrowing located?''.
For impression sentences, since we have already extracted disorders, we can generate case-level questions such as ``What disorder can be identified from this volume?'' for each case.
 
\begin{figure}[!htb]
\centering
\includegraphics[width=\linewidth]{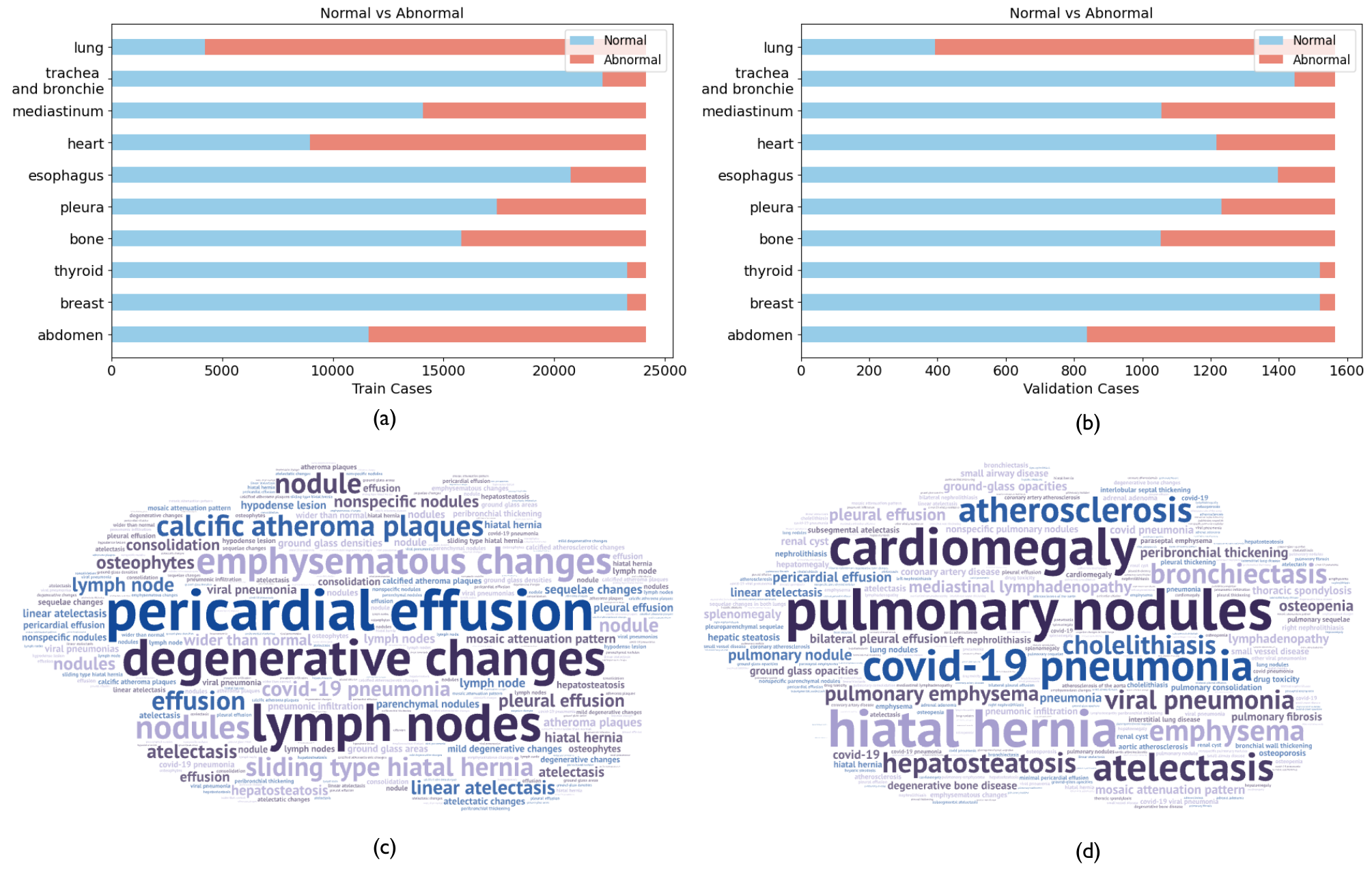}
\caption{(a) Distribution of Normal vs. Abnormal Cases: Training set analysis.
(b) Distribution of Normal vs. Abnormal Cases: Validation set analysis.
(c) Abnormalities Word Cloud: Visual summary of frequent abnormalities from findings.
(d) Disorders Word Cloud: Visual summary of frequent disorders from findings.}
\label{fig:region_abnormality}
\end{figure}

\noindent In summary, we have successfully generated 1.3M grounded Visual Question Answering (VQA) pairs for training and 85k for validation, along with 24,128 case-level visual question answering pairs for training and 1,564 for validation. Detailed counts for each type of VQA pair in both the training and validation sets are provided in Table~\ref{tab:datatype}.

\subsection*{Dataset Analysis}

% \subsubsection*{Abnormality Analysis} 
In this section, we analyze the abnormalities of the proposed dataset. The proposed hierarchically structured abnormality tree for each case enables us to systematically extract and analyze the abnormal findings.
First, we calculate the normal-to-abnormal case numbers for both the training set and the validation set, as shown in Fig.~\ref{fig:region_abnormality}.
Second, based on the identified and extracted abnormalities from all anatomical regions, we proceed to visualize these findings using a word cloud. 
The word cloud of abnormalities is presented in Fig.~\ref{fig:region_abnormality}.

\section*{Conclusion}
In this paper, we develop an automated pipeline for generating grounded datasets and introduce \OURDATA, a comprehensive, large-scale, region-guided 3D chest CT interpretation dataset based on CT-RATE. 
\OURDATA include \textbf{197} organ-level segmentation masks, \textbf{665 K}  multi-granularity grounded reports, and \textbf{1.3 M} grounded VQA pairs. We anticipate that \OURDATA will significantly advance multimodal medical AI models, enabling them to generate texts based on segmentation regions, thus enhancing interpretability and patient care. We will release all segmentation masks, grounded reports, and VQA pairs to support future research in this field.

 \clearpage
\bibliography{sample}

\clearpage 
\appendix

\section{Case Examples}
\label{sec:supple_case}

\begin{figure}[htb]
\centering
\includegraphics[width=\linewidth]{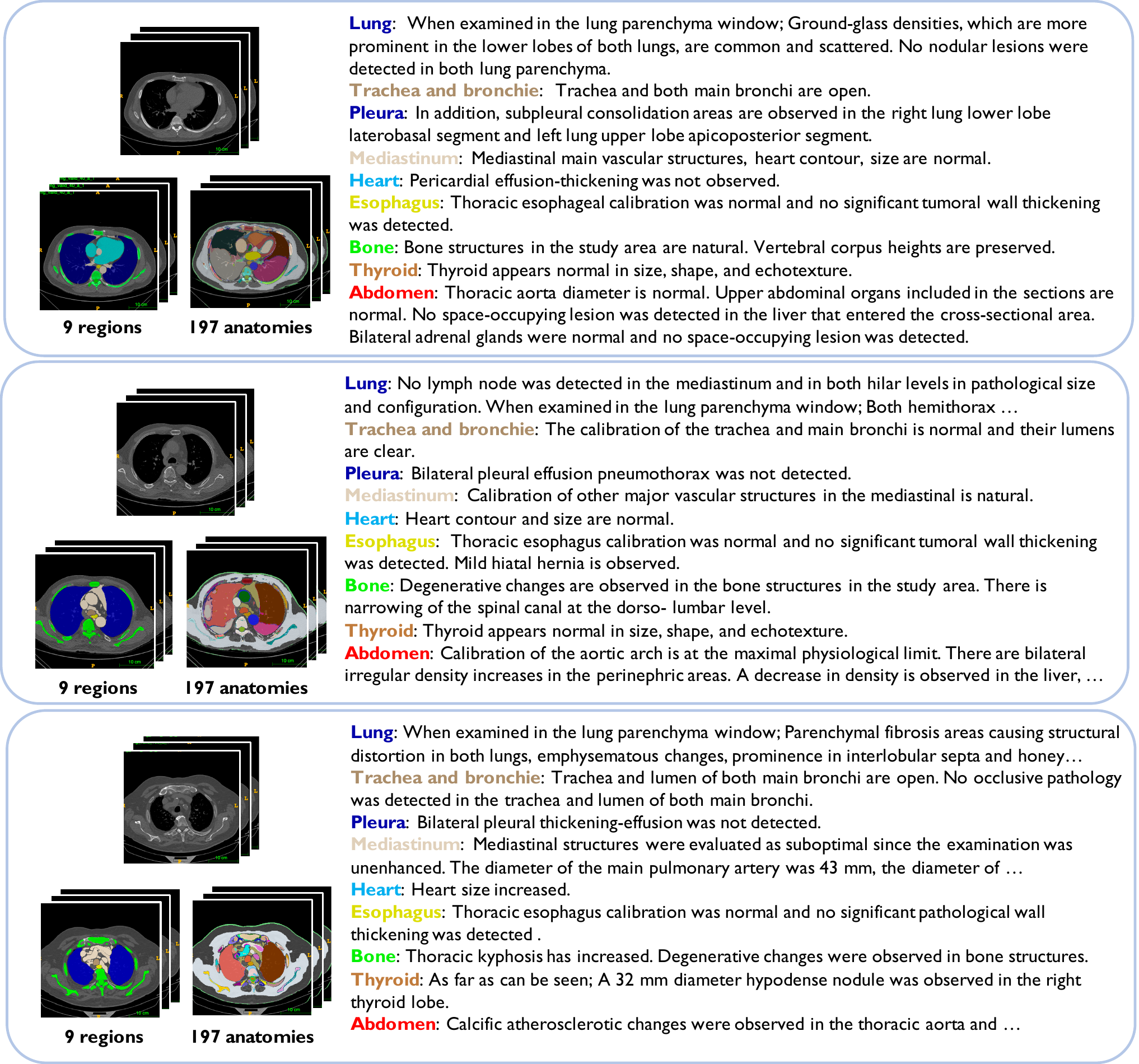}
\caption{Case examples of segmentation masks and structured reports.}
\label{fig:supple_case}
\end{figure}

\clearpage

\section{Hierarchy of anatomical regions}
\label{sec:hierarchy_anatomical}

\vspace{10pt}
\dirtree{%
    .1 \textbf{Chest CT Volume}.
    .2 \textbf{\textit{Lung}}.
    .3 Lung.
    .4 Left Lung.
    .5 Left Lung Lower Lobe.
    .5 Left Lung Upper Lobe.
    .4 Right Lung.
    .5 Right Lung Lower Lobe.
    .5 Right Lung Middle Lobe.
    .5 Right Lung Upper Lobe.
    .4 Lung Lower Lobe.
    .5 Left Lung Lower Lobe.
    .5 Right Lung Lower Lobe.
    .4 Lung Upper Lobe.
    .5 Left Lung Upper Lobe.
    .5 Right Lung Upper Lobe.
    .2 \textbf{\textit{Trachea and Bronchie}}.
    .3 Trachea.
    .3 Bronchie.
    .2 \textbf{\textit{Mediastinum}}.
    .3 Brachiocephalic Trunk.
    .3 Brachiocephalic Vein.
    .4 Left Brachiocephalic Vein.
    .4 Right Brachiocephalic Vein.
    .3 Superior Vena Cava.
    .3 Aorta.
    .3 Pulmonary Artery.
    .3 Thymus.
    .3 Pulmonary Vein.
    .3 Mediastinal Tissue.
    .3 Subclavian Artery.
    .4 Left Subclavian Artery.
    .4 Right Subclavian Artery.
    .2 \textbf{\textit{Heart}}.
    .3 Heart.
    .4 Heart Atrium.
    .5 Left Heart Atrium.
    .5 Right Heart Atrium.
    .4 Heart Ventricle.
    .5 Left Heart Ventricle.
    .5 Right Heart Ventricle.
    .4 Heart Ascending Aorta.
    .4 Heart Tissue.
    .5 Myocardium.
    .4 Left Auricle of Heart.
    .2 \textbf{\textit{Esophagus}}.
    .3 Esophagus.
    .4 Cervical Esophagus.
    .4 Cricopharyngeal Inlet.
    .2 \textbf{\textit{Pleura}}.
    .2 \textbf{\textit{Bone}}.
    .3 Bone.
    .4 Spinal Cord.
    .4 Spinal Canal.
    .4 Vertebrae.
    .5 Cervical Vertebrae.
    .6 Cervical Vertebrae 1 (C1).
    .6 Cervical Vertebrae 2 (C2).
    .6 Cervical Vertebrae 3 (C3).
    .6 Cervical Vertebrae 4 (C4).
    .6 Cervical Vertebrae 5 (C5).
    .6 Cervical Vertebrae 6 (C6).
    .6 Cervical Vertebrae 7 (C7).
    .5 Thoracic Vertebrae.
    .6 Thoracic Vertebrae 1 (T1).
    .6 Thoracic Vertebrae 2 (T2).
    .6 Thoracic Vertebrae 3 (T3).
    .6 Thoracic Vertebrae 4 (T4).
    .6 Thoracic Vertebrae 5 (T5).
    .6 Thoracic Vertebrae 6 (T6).
    .6 Thoracic Vertebrae 7 (T7).
    .6 Thoracic Vertebrae 8 (T8).
    .6 Thoracic Vertebrae 9 (T9).
    .6 Thoracic Vertebrae 10 (T10).
    .6 Thoracic Vertebrae 11 (T11).
    .6 Thoracic Vertebrae 12 (T12).
    .5 Lumbar Vertebrae.
    .6 Lumbar Vertebrae 1 (L1).
    .6 Lumbar Vertebrae 2 (L2).
    .6 Lumbar Vertebrae 3 (L3).
    .6 Lumbar Vertebrae 4 (L4).
    .6 Lumbar Vertebrae 5 (L5).
    .6 Lumbar Vertebrae 6 (L6).
    .5 Sacral Vertebrae 1 (S1).
    .4 Clavicle.
    .5 Left Clavicle.
    .5 Right Clavicle.
    .4 Scapula.
    .5 Left Scapula.
    .5 Right Scapula.
    .4 Humerus.
    .5 Left Humerus.
    .5 Right Humerus.
    .4 Femur.
    .5 Left Femur.
    .5 Right Femur.
    .4 Head of Femur.
    .5 Left Head of Femur.
    .5 Right Head of Femur.
    .4 Rib.
    .5 Left Rib.
    .6 Left Rib 1.
    .6 Left Rib 2.
    .6 Left Rib 3.
    .6 Left Rib 4.
    .6 Left Rib 5.
    .6 Left Rib 6.
    .6 Left Rib 7.
    .6 Left Rib 8.
    .6 Left Rib 9.
    .6 Left Rib 10.
    .6 Left Rib 11.
    .6 Left Rib 12.
    .5 Right Rib.
    .6 Right Rib 1.
    .6 Right Rib 2.
    .6 Right Rib 3.
    .6 Right Rib 4.
    .6 Right Rib 5.
    .6 Right Rib 6.
    .6 Right Rib 7.
    .6 Right Rib 8.
    .6 Right Rib 9.
    .6 Right Rib 10.
    .6 Right Rib 11.
    .6 Right Rib 12.
    .4 Rib 1.
    .5 Left Rib 1.
    .5 Right Rib 1.
    .4 Rib 2.
    .5 Left Rib 2.
    .5 Right Rib 2.
    .4 Rib 3.
    .5 Left Rib 3.
    .5 Right Rib 3.
    .4 Rib 4.
    .5 Left Rib 4.
    .5 Right Rib 4.
    .4 Rib 5.
    .5 Left Rib 5.
    .5 Right Rib 5.
    .4 Rib 6.
    .5 Left Rib 6.
    .5 Right Rib 6.
    .4 Rib 7.
    .5 Left Rib 7.
    .5 Right Rib 7.
    .4 Rib 8.
    .5 Left Rib 8.
    .5 Right Rib 8.
    .4 Rib 9.
    .5 Left Rib 9.
    .5 Right Rib 9.
    .4 Rib 10.
    .5 Left Rib 10.
    .5 Right Rib 10.
    .4 Rib 11.
    .5 Left Rib 11.
    .5 Right Rib 11.
    .4 Rib 12.
    .5 Left Rib 12.
    .5 Right Rib 12.
    .4 Rib Cartilage.
    .4 Costal Cartilage.
    .4 Sternum.
    .5 Manubrium of Sternum.
    .4 Eustachian Tube Bone.
    .5 Left Eustachian Tube Bone.
    .5 Right Eustachian Tube Bone.
    .2 \textbf{\textit{Thyroid}}.
    .3 Thyroid.
    .4 Thyroid Gland.
    .5 Left Thyroid.
    .5 Right Thyroid.
    .2 \textbf{\textit{Breast}}.
    .3 Left Breast.
    .3 Right Breast.
    .2 \textbf{\textit{Abdomen}}.
    .3 Abdomen.
    .4 Abdominal Tissue.
    .4 Adrenal Gland.
    .5 Left Adrenal Gland.
    .5 Right Adrenal Gland.
    .4 Aorta.
    .4 Colon.
    .4 Duodenum.
    .4 Gallbladder.
    .4 Intestine.
    .5 Small Bowel.
    .4 Kidney.
    .5 Left Kidney.
    .5 Right Kidney.
    .4 Liver.
    .5 Left Lobe of Liver.
    .6 Left Lateral Inferior Segment of Liver.
    .6 Left Lateral Superior Segment of Liver.
    .6 Left Medial Segment of Liver.
    .5 Right Lobe of Liver.
    .6 Right Anterior Inferior Segment of Liver.
    .6 Right Anterior Superior Segment of Liver.
    .6 Right Posterior Inferior Segment of Liver.
    .6 Right Posterior Superior Segment of Liver.
    .4 Liver Vessel.
    .4 Caudate Lobe.
    .4 Pancreas.
    .4 Portal Vein and Splenic Vein.
    .4 Rectum.
    .4 Renal Artery.
    .4 Renal Vein.
    .4 Spleen.
    .4 Stomach.
    .4 Celiac Trunk.
    .2 \textbf{\textit{Others}}.
    .3 Thoracic Cavity.
    .3 Prostate.
    .3 Urinary Bladder.
    .3 Carotid Artery.
    .4 Common Carotid Artery.
    .4 Internal Carotid Artery.
    .4 Left Carotid Artery.
    .5 Left Common Carotid Artery.
    .5 Left Internal Carotid Artery.
    .4 Right Carotid Artery.
    .5 Right Common Carotid Artery.
    .5 Right Internal Carotid Artery.
    .3 Iliac Artery.
    .3 Iliac Vena.
    .3 Iliac Vein.
    .3 Left Iliac Artery.
    .3 Left Iliac Vena.
    .3 Right Iliac Artery.
    .3 Right Iliac Vena.
    .3 Inferior Vena Cava.
    .3 Internal Jugular Vein.
    .3 Larynx.
    .4 Larynx Glottis.
    .4 Larynx Supraglottis.
    .3 Muscle.
}

\clearpage 

\section{Question template}

\subsubsection*{Abnormality}
\begin{enumerate}
    \item What are the abnormalities in the \{region\}?
    \item What abnormalities are present in the \{region\}?
    \item What types of abnormality are visible in the \{region\}?
    \item What type of abnormality is visible in the \{region\}?
    \item What kind of abnormalities are observed in the \{region\}?
    \item What kinds of abnormalities can be identified in the \{region\}?
    \item What specific abnormalities are detected in the \{region\}?
    \item What specific types of abnormalities are evident in the \{region\}?
    \item What types of abnormalities are evident upon examination of the \{region\}?
    \item What types of abnormalities can be seen in the \{region\}?
    \item What are the anomalies in the \{region\}?
    \item What anomalies are present in the \{region\}?
    \item What types of anomalies are visible in the \{region\}?
    \item What type of anomalies is visible in the \{region\}?
    \item What kind of anomalies are observed in the \{region\}?
    \item What kinds of anomalies can be identified in the \{region\}?
    \item What specific anomalies are detected in the \{region\}?
    \item What specific types of anomalies are evident in the \{region\}?
    \item What types of anomalies are evident upon examination of the \{region\}?
    \item What types of anomalies can be seen in the \{region\}?
    \item What are the abnormal findings in the \{region\}?
    \item What abnormal findings are present in the \{region\}?
    \item What types of abnormal findings are visible in the \{region\}?
    \item What type of abnormal findings is visible in the \{region\}?
    \item What kind of abnormal findings are observed in the \{region\}?
    \item What kinds of abnormal findings can be identified in the \{region\}?
    \item What specific abnormal findings are detected in the \{region\}?
    \item What specific types of abnormal findings are evident in the \{region\}?
    \item What types of abnormal findings are evident upon examination of the \{region\}?
    \item What types of abnormal findings can be seen in the \{region\}?
    \item What are the irregular findings in the \{region\}?
    \item What irregular findings are present in the \{region\}?
    \item What types of irregular findings are visible in the \{region\}?
    \item What type of irregular findings is visible in the \{region\}?
    \item What kind of irregular findings are observed in the \{region\}?
    \item What kinds of irregular findings can be identified in the \{region\}?
    \item What specific irregular findings are detected in the \{region\}?
    \item What specific types of irregular findings are evident in the \{region\}?
    \item What types of irregular findings are evident upon examination of the \{region\}?
    \item What types of irregular findings can be seen in the \{region\}?    
    \item What are the irregularities in the \{region\}?
    \item What irregularities are present in the \{region\}?
    \item What types of irregularities are visible in the \{region\}?
    \item What type of irregularities is visible in the \{region\}?
    \item What kind of irregularities are observed in the \{region\}?
    \item What kinds of irregularities can be identified in the \{region\}?
    \item What specific irregularities are detected in the \{region\}?
    \item What specific types of irregularities are evident in the \{region\}?
    \item What types of irregularities are evident upon examination of the \{region\}?
    \item What types of irregularities can be seen in the \{region\}?
\end{enumerate}

\subsubsection*{Presence}
\begin{enumerate}
    \item Can \{abnormality\} be identified in the \{region\}?
    \item Can \{abnormality\} be observed in the \{region\}?
    \item Can \{abnormality\} be detected in the \{region\}?
    \item Can \{abnormality\} be seen in the \{region\}?
    \item Can \{abnormality\} be founded in the \{region\}?
    \item Can \{abnormality\} be recognized in the \{region\}?
    \item Can we detect any signs of \{abnormality\} in the \{region\}?
    \item Can we observe any signs of \{abnormality\} in the \{region\}?
    \item Can we recognize any signs of \{abnormality\} in the \{region\}?
    \item Can we see \{abnormality\} in the \{region\}?
    \item Can we find \{abnormality\} in the \{region\}?
    \item Can we detect \{abnormality\} in the \{region\}?
    \item Can we observe \{abnormality\} in the \{region\}?
    \item Can we identify \{abnormality\} in the \{region\}?
    \item Is there any sign of \{abnormality\} in the \{region\}?
    \item Is there any indication of \{abnormality\} in the \{region\}?
    \item Is there any evidence of \{abnormality\} in the \{region\}?
    \item Is there any suggestion of \{abnormality\} in the \{region\}?
    \item Is there a clear sign of \{abnormality\} in the \{region\}?
    \item Is there a clear indication of \{abnormality\} in the \{region\}?
    \item Is there a clear evidence of \{abnormality\} in the \{region\}?
    \item Is there a clear suggestion of \{abnormality\} in the \{region\}?
    \item Is \{abnormality\} visibly present in the \{region\}?
    \item Is \{abnormality\} clearly visible in the \{region\}?
    \item Is there any visual evidence suggesting \{abnormality\} in the \{region\}?
    \item Is there any indication of \{abnormality\} upon examination of the \{region\}?
    \item Is there any indication of \{abnormality\} in the \{region\}?
    \item Is there visual evidence of \{abnormality\} in the \{region\} on this scan?
    \item Are there any visible indications of \{abnormality\} in this \{region\}?
    \item Are there any visible cues indicating \{abnormality\} in the \{region\}?
    \item Are there any visible indicators of \{abnormality\} in the \{region\}?
    \item Are there any clear indications of \{abnormality\} in this \{region\}?
    \item Are there any clear cues indicating \{abnormality\} in the \{region\}?
    \item Are there any clear indicators of \{abnormality\} in the \{region\}?
    \item Are there any indications of \{abnormality\} in this \{region\}?
    \item Are there any cues indicating \{abnormality\} in the \{region\}?
    \item Are there any indicators of \{abnormality\} in the \{region\}?
    \item Are there any observable signs of \{abnormality\} in the \{region\}?
    \item Are there any signs of \{abnormality\} in the \{region\}?
    \item Are there any features of \{abnormality\} in the \{region\}?
    \item Does the \{region\} show the presence of \{abnormality\}?
    \item Does the \{region\} show the existence of \{abnormality\}?
    \item Does the image suggest the presence of \{abnormality\} in the \{region\}?
    \item Does the image suggest the existence of \{abnormality\} in the \{region\}?
    \item Does the \{region\} exhibit any evidence of \{abnormality\}?
    \item Does the \{region\} display any features suggestive of \{abnormality\}?
    \item Does the \{region\} display any characteristics suggestive of \{abnormality\}?
    \item Does the \{region\} exhibit any characteristics indicative of \{abnormality\}?
    \item Does the \{region\} exhibit any features indicative of \{abnormality\}?
    \item Does the visual features suggest the presence of \{abnormality\} in the \{region\}?
\end{enumerate}

\subsubsection*{Location}
\begin{enumerate}
    \item Where is the \{abnormality\} located in the image?
    \item Where can the \{abnormality\} be found within the image?
    \item Where in the image is the \{abnormality\} located?
    \item Where in the image is the \{abnormality\} localized?
    \item Where in the image can the \{abnormality\} be found?
    \item Where in the image does the \{abnormality\} appear?
    \item Where in the image does the \{abnormality\} locate?
    \item Where in the image does the \{abnormality\} locate?
    \item Where specifically within the image is the \{abnormality\} located?
    \item Where exactly within the image is the \{abnormality\} located?
    \item Where exactly is the \{abnormality\} located in the image?
    \item Where specifically is the \{abnormality\} located in the image?
    \item Where exactly within the image is the \{abnormality\} localized?
    \item Where specifically within the image is the \{abnormality\} localized?
    \item Where within the image can the \{abnormality\} be precisely located?
    \item Where exactly within the image does the \{abnormality\} present?
    \item Where within the image does the \{abnormality\} specifically present?
    \item Where in the image does the \{abnormality\} appear?
    \item What is the location of the \{abnormality\} in the image?
    \item What is the precise location of the \{abnormality\} in the image?
    \item What is the specific location of the \{abnormality\} within the image?
    \item What is the precise region of the \{abnormality\} in the image?
    \item What is the specific region of the \{abnormality\} within the image?
    \item What particular region within the image does the \{abnormality\} occupy?
    \item What particular location within the image does the \{abnormality\} occupy?
    \item What specific location within the image does the \{abnormality\} occupy?
    \item What specific region within the image does the \{abnormality\} occupy?
    \item What specific area of the image does the \{abnormality\} occupy?
    \item What specific region of the image does the \{abnormality\} appear?
    \item What specific spot within the image contains the \{abnormality\}?
    \item What particular region of the image is affected by the \{abnormality\}?
    \item What specific area within the image is impacted by the \{abnormality\}?
    \item What specific region within the image is impacted by the \{abnormality\}?
    \item What specific location within the image is impacted by the \{abnormality\}?
    \item What particular region within the image is affected by the \{abnormality\}?
    \item What particular area within the image is affected by the \{abnormality\}?
    \item What particular location within the image is affected by the \{abnormality\}?
    \item What specific region within the image does the \{abnormality\} affect?
    \item What specific area within the image does the \{abnormality\} affect?
    \item What specific location within the image does the \{abnormality\} affect?
    \item What specific location within the image does the \{abnormality\} appear?
    \item What specific region within the image does the \{abnormality\} appear?
    \item What specific area within the image does the \{abnormality\} appear?
    \item What particular spot within the image does the \{abnormality\} present?
    \item What particular area within the image does the \{abnormality\} present?
    \item What particular region within the image does the \{abnormality\} present?
    \item What particular location within the image does the \{abnormality\} present?
    \item What specific area within the image does the \{abnormality\} occur?
    \item What specific location within the image does the \{abnormality\} occur?
    \item What specific region within the image does the \{abnormality\} occur?
\end{enumerate}

\subsubsection*{Size}
\begin{enumerate}
    \item What is the approximate size of the \{abnormality\} in the \{region\}?
    \item What is the approximate scale of the \{abnormality\} in the \{region\}?
    \item What is the approximate size range of the \{abnormality\} in the \{region\}?
    \item What is the approximate magnitude of the \{abnormality\} in the \{region\}?
    \item What is the approximate dimension of the \{abnormality\} in the \{region\}?
    \item What is the approximate measurement of the \{abnormality\} in the \{region\}?
    \item What is the estimated size of the \{abnormality\} in the \{region\}?
    \item What is the estimated scale of the \{abnormality\} in the \{region\}
    \item What is the estimated size range of the \{abnormality\} in the \{region\}?
    \item What is the estimated magnitude of the \{abnormality\} in the \{region\}?
    \item What is the estimated dimension of the \{abnormality\} in the \{region\}?
    \item What is the estimated measurement of the \{abnormality\} in the \{region\}?
    \item What is the size assessment of the \{abnormality\} in the \{region\}?
    \item What is the measurement of the \{abnormality\} in the \{region\}?
    \item What is the scale of the \{abnormality\} in the \{region\}?
    \item What is the size of the \{abnormality\} in the \{region\}?
    \item What is the size range of the \{abnormality\} in the \{region\}?
    \item What is the magnitude of the \{abnormality\} in the \{region\}?
    \item What is the dimension of the \{abnormality\} in the \{region\}?
    \item What is the overall size of the \{abnormality\} in the \{region\}?
    \item What is the overall scale of the \{abnormality\} in the \{region\}?
    \item What is the overall measurement of the \{abnormality\} in the \{region\}?
    \item What is the overall size range of the \{abnormality\} in the \{region\}?
    \item What is the overall magnitude of the \{abnormality\} in the \{region\}?
    \item What is the overall dimension of the \{abnormality\} in the \{region\}?
    \item What is the scale of the \{abnormality\} detected in the \{region\}?
    \item What is the size of the \{abnormality\} detected in the \{region\}?
    \item What is the size range of the \{abnormality\} detected in the \{region\}?
    \item What is the measurement of the \{abnormality\} detected in the \{region\}?
    \item What is the magnitude of the \{abnormality\} detected in the \{region\}?
    \item What is the dimension of the \{abnormality\} detected in the \{region\}?
    \item What is the scale of the \{abnormality\} appeared in the \{region\}?
    \item What is the size of the \{abnormality\} appeared in the \{region\}?
    \item What is the size range of the \{abnormality\} appeared in the \{region\}?
    \item What is the magnitude of the \{abnormality\} appeared in the \{region\}?
    \item What is the dimension of the \{abnormality\} appeared in the \{region\}?
    \item What is the measurement of the \{abnormality\} appeared in the \{region\}?
    \item How large is the affected \{abnormality\} area in the \{region\}?
    \item How large is the observed \{abnormality\} area in the \{region\}?
    \item How large is the \{abnormality\} observed in the \{region\}?
    \item How large is the \{abnormality\} in the \{region\}?
    \item How large does the \{abnormality\} appear in the \{region\}?
    \item How large does the \{abnormality\} appear to be in the \{region\}?
    \item How large is the \{abnormality\} area in the \{region\}?
    \item How large is the affected \{abnormality\} area in the \{region\}?
    \item How big is the affected \{abnormality\} area in the \{region\}?
    \item How big is the observed \{abnormality\} area in the \{region\}?
    \item How big is the \{abnormality\} observed in the \{region\}?
    \item How big is the \{abnormality\} in the \{region\}?
\end{enumerate}

\end{document}